\documentclass[runningheads]{llncs}
\usepackage{graphicx}
\usepackage{tabularx}
\usepackage{hyperref}

\begin{document}
\title{Structured Summarization of \\ Academic Publications}
%
%
\author{Alexios Gidiotis \and
Grigorios Tsoumakas}
\authorrunning{A. Gidiotis and G. Tsoumakas}
%
\institute{School of Informatics, Aristotle University of Thessaloniki}
\maketitle              
\begin{abstract}
We propose SUSIE, a novel summarization method that can work with state-of-the-art summarization models in order to produce structured scientific summaries for academic articles. We also created PMC-SA, a new dataset of academic publications, suitable for the task of structured summarization with neural networks. We apply SUSIE combined with three different summarization models on the new PMC-SA dataset and we show that the proposed method improves the performance of all models by as much as 4 ROUGE points.
\keywords{Text summarization \and natural language processing \and deep learning.}
\end{abstract}
\section{Introduction}
Having informative summaries of scientific articles is crucial for dealing with the avalanche of academic publications in our times. Such summaries would allow researchers to quickly and accurately screen retrieved articles for relevance to their interests. More importantly, such summaries would lead to high quality indexing of the articles by (academic) search engines, leading to more relevant academic search results.     


Currently, the role of such summaries is played by the abstracts produced by the authors of the articles. However, authors usually include in the abstract only the contributions and information of the paper that they consider important and ignore others that might be equally important to the scientific community \cite{Elkiss2008BlindArticle}. 




A solution to the above problem would be to employ state-of-the-art summarization approaches \cite{See2017GetNetworks,Paulus2017ASummarization,Dangovski2019RotationalApplications,Celikyilmaz2018DeepSummarization}, in order to automatically create short informative summaries of the articles to replace and/or accompany author abstracts for machine indexing and human inspection. These approaches however, have focused on the summarization of newswire articles while academic articles exhibit several differences and pose major challenges compared to news articles. 

First of all, news articles are much shorter than scientific articles and the news headlines that serve as summaries are much shorter than scientific abstracts. Secondly, scientific articles usually include several different key points that are scattered throughout the paper and need to be accurately included in a summary. These problems make it difficult to use summarization models that achieve state-of-the-art performance on newswire datasets for the summarization of academic articles.

We propose SUSIE (StrUctured SummarIzEr), a novel training method that allows us to effectively train existing summarization models on academic articles that have structured abstracts. Our method uses the XML structure of the articles and abstracts in order to split each article into multiple training examples and train summarization models that learn to summarize each section separately. We call such a task {\em structured summarization}. We further contribute a novel dataset consisting of open access PubMed Central articles along with their structured abstracts. SUSIE can easily be combined with different summarization models in order to address the problem of long articles and has been found to improve the performance of state-of-the-art summarization models by 4 ROUGE points.

We also created {\em PMC-SA} (PMC Structured Abstracts), a novel dataset that consists of academic articles from the biomedical domain The articles for this dataset were collected from the {\em PubMed Central Open Access} (PMC-OA) repository and follow the {\em IMRD} ((Introduction, Methods, Results, Discussion) structure. The abstracts in this dataset are also structured in a similar manner and each section of the full text can be paired with the corresponding section of the abstract.

\section{Related Work}

\subsection{Summarization Methods}
\label{sec:seq2seq_models}

State-of-the-art summarization methods use recurrent neural networks (RNNs) with the encoder-decoder architecture (or sequence-to-sequence architecture). These  methods usually treat the whole source text as an input sequence, encode it into their hidden state and generate a complete summary from that hidden state. 

Strong results have been achieved by such models when combined with an {\em attention} mechanism \cite{Nallapati2016AbstractiveBeyond,Chopra2016AbstractiveNetworks,Rush2015ASummarization}. Adding a {\em pointer-generator} mechanism has been shown to further improve results  \cite{See2017GetNetworks}. The pointer-generator mechanism gives the model the ability to copy important words from the source text in addition to generating words from a predefined vocabulary. Adding a {\em coverage} mechanism has been shown to lead to even better results. \cite{See2017GetNetworks}. The coverage mechanism prevents the model from repeating itself, which is a common problem with sequence-to-sequence models. In recent work, various approaches utilized reinforcement learning and policy gradient to further improve the performance of baseline models \cite{Paulus2017ASummarization,Celikyilmaz2018DeepSummarization}. Finally, \cite{Dangovski2019RotationalApplications} replaced the LSTM cells in the model of \cite{See2017GetNetworks} with a different type of RNN unit in order to overcome fundamental limitations of LSTM cells.
 
\subsection{Summarization Datasets}
\label{sec:other_data}
Most of the summarization datasets that are found in the literature such as {\em Newsroom}  \cite{Grusky2018Newsroom:Strategies}, {\em Gigaword}  \cite{Napoles2012AnnotatedGigaword} and {\em CNN / Daily Mail}  \cite{Hermann2015TeachingComprehend} are focused on newswire articles.
The average article lengths are relatively small and range from 50 words (Gigaword) to a few hundred words (CNN / Daily Mail, Newsroom). The average summary lengths are also rather small and range from a single sentence (Gigaword, Newsroom) to a few sentences (CNN / Daily Mail). 

TAC 2014 (Text Analysis Conference 2014) is the only dataset that focuses on (biomedical) academic articles. The articles have an average of 9,759 words and the summaries an average of 235 words. However, as it consists of just 20 articles, it is not useful for training complex neural network summarization models. 


\section{Summarizing Academic Papers}
\subsection{Flat Abstract Summarization}
\label{sec:flat_sum}

A simple approach to summarizing academic papers would be to train sequence-to-sequence models using the full text of the article as source input and the abstract as reference summary. However, sequence-to-sequence models face multiple difficulties when given long input texts. A very long input sequence requires the encoder RNN to run for a lot of time steps. This greatly increases the computational complexity of the forward pass. To make things worse, the training of the encoder on very long input sequences becomes increasingly difficult due to the computational complexity of the backward pass. The training becomes increasingly slower and in many cases the vanishing gradients prevent the model from learning useful information.

A solution to this problem would be to truncate very long sequences (more than 600 words), but this can result in serious information loss which would severely affect the quality of the produced summaries.

Even harder is the training of a decoder with very long output sequences. In this case, the computational complexity and memory requirements of the decoder make it pointless to try and train a model with very long reference summaries.

Another problem of this straightforward approach, is that the different parts of an academic paper are not equally important for the task of summarization. Sections like the {\em introduction} include core information for the summary, while others like the {\em experiments} are noisy and usually include little useful information.

\subsection{SUSIE}
\label{sec:susie}

SUSIE (StrUctured SummarIzEr) is a novel summarization method that exploits structured abstracts in order to address the aforementioned problems. 

Many academic articles, especially in the life sciences domain follow the typical {\em IMRD} structure with sections like {\em introduction}, {\em background}, {\em methods}, {\em results} and {\em  conclusion}. When the abstract of the article is structured it usually  includes similar sections too. We employed a very simple method that looks for specific keywords in the header of each section in order to annotate both the article and abstract sections. For example, sections that include keywords like {\em methods}, {\em method}, {\em techniques} and {\em methodology} in their header are annotated as \textit{methods}. Table \ref{section_tags} presents the different section types and the keywords associated with them. 

\begin{table}[t!]
\begin{center}
\setlength\tabcolsep{0.2cm}
\begin{tabular}{|r|l|}
\hline \textbf{section} & \textbf{keywords}\\\hline
introduction & introduction, case\\
literature & background, literature, related\\
methods & methods, method, techniques, methodology\\
results & result, results, experimental, experiments, experiment\\
discussion & discussion, limitations\\
conclusion & conclusion, conclusions, concluding\\
\hline
\end{tabular}
\end{center}
\caption{\label{section_tags} The different sections that we annotate and the keywords associated with them.}
\end{table}

Once the article and abstract sections are annotated, we pair each section of the full text with the corresponding section of the abstract and create one training example per section. We can then use one of the existing summarization methods and train a model for the summarization of single sections. Summarizing a single section of an article is a much easier task since the input and output sequences are a lot shorter and the information is more compact and focused on specific aspects of the article. In addition, section annotation allows us to filter out particular sections that are not useful for summarization.

At test time we extract the specified sections of the article and run the summarization model for each of them in order to produce section summaries. Then we combine those summaries in order to get the full summary of the article.


\section{PMC Structured Abstracts}

PubMed Central (PMC) is a free digital repository that archives publicly accessible full-text scholarly articles that have been published within the biomedical and life sciences journal literature. The PMC-SA (PMC Structured Abstracts) dataset was created from the open access subset of PMC, comprising approximately 2 million articles. We used the XML format downloaded from the PMC FTP server to create the dataset. Only the articles that have abstracts structured in sections were selected and included in the dataset. PMC-SA has a total of 712,911 full text articles along with their abstracts. The full texts of the articles have an average length of 2,514 words and are used as source texts for the summarization, while the abstracts have an average length of 260 words and are used as reference summaries. Code and instructions for the creation of the PMC-SA dataset will be made available online.\footnote{\url{https://github.com/AlexGidiotis/PMC-StructuredAbstracts-Dataset}} When compared with the existing datasets discussed in section \ref{sec:other_data} PMC-SA is clearly different in multiple ways. The articles and summaries are significantly longer compared to the different newswire datasets and this makes it a much harder task. Also, the new dataset is a lot larger than the TAC 2014 dataset which is the only other dataset that consists of academic publications. This makes it suitable for the training of state-of-the-art summarization models. 

We can easily apply SUSIE on PMC-SA since the XML format allows us to effectively split the full text and abstract into annotated sections. In table \ref{per_sec_exp_stats} we show detailed statistics about the source and abstract length for each section type.

\begin{table}[t!]
\begin{center}
\setlength\tabcolsep{0.2cm}
\begin{tabular}{|r|rr|rr|}
\hline & \multicolumn{2}{c|}{\textbf{Source length}} & \multicolumn{2}{c|}{\textbf{Abstract length}} \\ 
\textbf{section type} & \textbf{mean} & \textbf{std} & \textbf{mean} & \textbf{std}\\\hline
introduction & 570.26 & 381.40 & 58.25 & 41.00\\
methods & 1,133.32 & 638.90 & 80.26 & 38.98 \\
conclusion & 152.08 & 178.14 & 49.92 & 23.83 \\
\hline
\end{tabular}
\end{center}
\caption{\label{per_sec_exp_stats} Per section type number of words for the articles in the PMC-SA dataset.}
\end{table}

\section{Experiments}

\begin{table*}[t!]
\begin{center}
\setlength\tabcolsep{0.1cm}
\begin{tabular}{|r|cc|cc|cc|}
\hline  & \multicolumn{2}{c|}{\textbf{ROUGE-1 F1}} & \multicolumn{2}{c|}{\textbf{ROUGE-2 F1}} & \multicolumn{2}{c|}{\textbf{ROUGE-L F1}}\\
{\bf Model}  & {\bf Flat} & {\bf SUSIE} & {\bf Flat} & {\bf SUSIE} & {\bf Flat} & {\bf SUSIE} \\\hline
attention sequence-to-sequence & 0.2833 & \textbf{0.3341} & 0.1043 & \textbf{0.1261} & 0.2619 & \textbf{0.3026} \\
pointer-generator & 0.3020 & \textbf{0.3591} & 0.1020 & \textbf{0.1416} & 0.2726 & \textbf{0.3179}\\
pointer-generator + coverage & 0.3300 & \textbf{0.3716} & 0.1142 & \textbf{0.1466} & 0.2893 & \textbf{0.3296}\\
\hline
\end{tabular}
\end{center}
\caption{\label{exp_results} Experimental results. Best result per evaluation measure is highlighted in bold typeface.}
\end{table*}

As we mentioned, SUSIE can be combined with a number of different summarization models. In order to evaluate the effectiveness of SUSIE the three different summarization models that were described in section \ref{sec:seq2seq_models} are trained and evaluated on PMC-SA using both the flat abstract method from section \ref{sec:flat_sum} and SUSIE.


The training set has 641,994 articles, the validation set has 35,309 articles and the test set 10,111 articles. In all experiments we included for summarization only the {\em introduction}, {\em methods} and {\em conclusion} sections because we have found that these particular section selection gives us the best performing models. For the flat abstract method, the selected sections are concatenated and used as source input paired with the concatenation of the corresponding abstract sections as reference summary. For SUSIE one example is created for each of the selected sections with the corresponding abstract section as reference summary. In Tables \ref{exp_stats} we provide detailed statistics about the training data used in the two different methods.

\begin{table}[t!]
\begin{center}
\setlength\tabcolsep{0.2cm}
\begin{tabular}{|r|rr|}
\hline & \textbf{Flat} & \textbf{SUSIE}\\\hline
\# training articles & 641,994 & 641,994\\
\# training examples & 641,994 & 1,211,826\\
avg. source length (words)& 1,451 & 677\\
avg. summary length (words)& 260 & 130\\
\hline
\end{tabular}
\end{center}
\caption{\label{exp_stats} Statistics about the training sets for the two experiments. In the flat abstract experiment each training example is an article and the whole abstract is used as reference summary. With SUSIE we create an average of 2 examples per article. The source inputs are article sections and the corresponding abstract sections are the reference summaries.}
\end{table}

\subsection{Experimental Setup}

We used the implementation of the three models provided by \cite{See2017GetNetworks}\footnote{\url{https://github.com/abisee/pointer-generator}}. The hyperparameter setup used for the models is similar to that of \cite{See2017GetNetworks} and is detailed in the supplementary material. 




In order to speed up the training process we start off with highly truncated input and output sequences. In more detail, we begin with input and output sequences truncated to 50 and 10 words respectively and train until convergence. Then we gradually increase the input and output sequences up to 500 and 100 words respectively.

When using the flat abstract method, we truncate each section to $\frac{L}{n}$ words before concatenating them to get the input and output sequences, where $L$ is the required article length and $n$ is the number of extracted sections from this article. 

The truncation of an academic article to a total of 500 words is definitely going to result in some severe information loss but we deemed it necessary due to the difficulties described in section \ref{sec:flat_sum}. To get the coverage model we simply add the coverage mechanism to the converged pointer-generator model and continue training.



At test time, for the flat abstract method, we truncate each input section to $\frac{L}{n}$ with $L=500$ words and concatenate them to get an input sequence of 500 words. Then we run beam search for 120 decoding steps in order to generate a summary. For SUSIE each of the selected sections is truncated to 500 words before we run beam search for 120 decoding steps to get a summary for each one of them. Then we concatenate the individual summaries to get the summary of the full article.

\subsection{Results}

We evaluate the performance of all models with the ROUGE family of metrics~\cite{Lin2004Rouge:Summaries} using the {\em pyrouge} package\footnote{\url{https://pypi.org/project/pyrouge/0.1.3}}. In specific, we report F1 scores for ROUGE-1, ROUGE-2 and ROUGE-L. ROUGE-1 and ROUGE-2 measure the overlap, in unigrams and bigrams respectively, between the generated and the reference summary. ROUGE-L measures the longest common subsequence overlap.

Table \ref{exp_results} presents the results of our experiments. We can see that the pointer-generator model achieves higher scores than the simple attention sequence-to-sequence and adding the coverage mechanism further improves those scores which is in line with the experiments of \cite{See2017GetNetworks}. 

We also notice that SUSIE improves the scores of the flat summarization approach for all three models by as much as 4 ROUGE points. The performance of the best model, pointer-generator with coverage, is improved by approximately 13\%, 28\% and 14\% in terms of ROUGE-1, ROUGE-2 and ROUGE-L F1 score respectively. It is clear that the flat approach suffers from information loss due to the truncation of the source input.  
In the appendix we illustrate the difference in the quality of the summaries produced by the two different methods by presenting generated examples for a real article. 

\section{Conclusion}

This work focused on the summarization of academic publications. We have shown that summarization models that perform well on smaller articles have difficulties when applied on longer articles with a lot of diverse information like academic articles. We proposed SUSIE, a novel approach that allowed us to successfully adapt existing summarization models to the task of structured summarization of academic articles. Also, we created PMC-SA, a new dataset of academic articles that is suitable for the training of summarization models using SUSIE. We found that training with SUSIE on the PMC-SA greatly improves the performance of summarization models and the quality of the generated summaries. 

\bibliographystyle{splncs04}
\bibliography{references}

\section{Appendix}
\small
Here we will provide an example of summaries generated by the best performing model, namely pointer-generator with coverage, for a sample article from the test set. We provide two summaries, one generated from a model trained with the flat method and another generated from a model trained with SUSIE. We also provide the original abstract of the article for reference.  One can find the original article with {\em PMCID PMC5051331} at the PMC website\footnote{\url{https://www.ncbi.nlm.nih.gov/pmc/articles/PMC5051331}}.

Comparing the two generated summaries, we can see that the one generated with SUSIE is superior to the flat one in terms of structure, readability and factual correctness. When compared with the original abstract, we can see that both summaries are not perfect but the one generated with SUSIE is in many cases acceptable. 

\subsection{Reference Summary}
\textbf{Objective} To examine the efficacy of psychological and psychosocial interventions for reductions in repeated self-harm. \\ 
\textbf{Design} We conducted a systematic review, meta-analysis and meta-regression to examine the efficacy of psychological and psychosocial interventions to reduce repeat self-harm in adults. We included a sensitivity analysis of studies with a low risk of bias for the meta-analysis. For the meta-regression, we examined whether the type, intensity (primary analyses) and other components of intervention or methodology (secondary analyses) modified the overall intervention effect. \\
\textbf{Data sources} A comprehensive search of medline, psycinfo and embase (from 1999 to june 2016) was performed. \\
\textbf{Eligibility criteria for selecting studies} Randomised controlled trials of psychological and psychosocial interventions for adult self-harm patients. \\
\textbf{Conclusions} Consideration of a psychological or psychosocial intervention over and above treatment as usual is worthwhile; with the public health benefits of ensuring that this practice is widely adopted potentially worth the investment. However, the specific type and nature of the intervention that should be delivered is not yet clear. Cognitive–behavioural therapy or interventions with an interpersonal focus and targeted on the precipitants to self-harm may be the best candidates on the current evidence. Further research is required.

\subsection{Flat P-Gen + Coverage}
Clinically treated non-fatal self-harm (we use the term self-harm henceforth) is common in terms of adverse outcomes such as repetition of self-harm, suicide and all-cause mortality; mental health morbidity, quality of life and physical, psychological and social functioning, mental health morbidity, quality of life and physical, psychological and social functioning. It is also costly in terms of immediate and ongoing treatment of self-harm of 16.3\% after 1 year and 7\% after 9 years. A more recent review of 177 studies from western and non-western countries indicated little change in these estimates, with reported repetition of self-harm of 16\%, while the suicide rate was 1.6

\subsection{SUSIE P-Gen + Coverage}
\textbf{Objective} To evaluate the efficacy of psychological and pharmacological interventions for reducing repetition of self-harm in unselected populations, but again highlight the poor quality of the evidence base in pooling data. In addition to these cochrane reviews, a number of trials have been published and several systematic reviews produced that aim to highlight what interventions are most often efficacious interventions (cbt) (and not problem-solving therapy). \\
\textbf{Methods} We searched medline, embase, and the cochrane central register of controlled controlled trials (rct) published up to february 2016 to identify randomized controlled trials evaluating the efficacy of psychological and psychosocial interventions to reduce repeat self-harm, (primary outcome) and to reduce suicidal ideation, depression and hopelessness (secondary outcomes) using meta-analysis; and (2) examine whether the type, intensity or other specific components of the interventions, or study methodology, modify the pooled intervention effect using meta-regression analysis. \\
\textbf{Conclusions} our study is consistent with the updated cochrane review, which in contrast to the original version showed no support for problem-focused (predominantly problem-solving therapy) interventions but a significant effect of interventions. Our study has shown that psychological or psychosocial interventions are effective overall, with cbt and psychodynamic interpersonal therapy currently the most promising for implementation.

\end{document}